%% file: Main_paper.tex
\documentclass[runningheads]{llncs}
\usepackage{geometry} 
\usepackage[numbers, sort&compress, sectionbib]{natbib} 
\usepackage{amsfonts}  
\usepackage{bm} 
\usepackage{enumitem} 

\usepackage{booktabs}       
\usepackage{multirow} 
\usepackage{graphicx}       
\usepackage{subfigure}      
\usepackage{array}
\newcolumntype{P}[1]{>{\centering\arraybackslash}p{#1}} 
\usepackage[dvipsnames]{xcolor} 
\usepackage{amssymb}
\usepackage{adjustbox}
\usepackage{tikz}  
\usetikzlibrary{patterns} 

\usepackage[belowskip=25pt,aboveskip=25pt]{caption} 
\usepackage[font=small]{caption} 
\usepackage[labelfont=bf]{caption} 

\usepackage{etoolbox}
\usepackage{xpatch}
\patchcmd{\thebibliography}{\chapter*}{\section*}{}{} 
\usepackage{hyperref} 

\newcommand{\appendixnumberline}[1]{Appendix\space}
\let\oldappendix\appendix
\makeatletter
\renewcommand{\appendix}{%
  \addtocontents{toc}{\let\protect\numberline\protect\appendixnumberline}%
  \renewcommand{\@seccntformat}[1]{Appendix~\csname the##1\endcsname\quad}%
  \oldappendix
}
\makeatother

\begin{document}

\title{Personalized Anomaly Detection in PPG Data using Representation Learning and Biometric Identification}

\author{Ramin Ghorbani\inst{1}\thanks{Corresponding author~(\email{r.ghorbani@tudelft.nl})}\and
Marcel J.T. Reinders\inst{1}\and
David M.J. Tax\inst{1}}

\institute{Pattern Recognition and Bioinformatics group, Delft University of Technology, Delft, Netherlands}

\maketitle              
\pagestyle{plain}

\begin{abstract}
Photoplethysmography (PPG) signals, typically acquired from wearable devices, hold significant potential for continuous fitness-health monitoring. In particular, heart conditions that manifest in rare and subtle deviating heart patterns may be interesting. However, robust and reliable anomaly detection within these data remains a challenge due to the scarcity of labeled data and high inter-subject variability. This paper introduces a two-stage framework leveraging representation learning and personalization to improve anomaly detection performance in PPG data. The proposed framework first employs representation learning to transform the original PPG signals into a more discriminative and compact representation. We then apply three different unsupervised anomaly detection methods for movement detection and biometric identification. We validate our approach using two different datasets in both generalized and personalized scenarios. The results show that representation learning significantly improves anomaly detection performance while reducing the high inter-subject variability. Personalized models further enhance anomaly detection performance, underscoring the role of personalization in PPG-based fitness-health monitoring systems. The results from biometric identification show that it's easier to distinguish a new user from one intended authorized user than from a group of users. Overall, this study provides evidence of the effectiveness of representation learning and personalization for anomaly detection in PPG data. 

\end{abstract}
\section{Introduction}
Photoplethysmography (PPG) data is a non-invasive, low-cost, optical physiological signal that measures the volume of blood flowing through the blood vessels and can be measured by a variety of wearable devices and smartwatches~\cite{r3}. PPG data enables remote health monitoring and fitness tracking, which presents opportunities for identifying unusual patterns in the user data that may indicate potential health issues, like abnormal heart rate or irregular movement patterns~\cite{r4}.

The effectiveness of detecting anomalies largely depends on the availability of enough labeled data. Supervised machine learning methods, such as k-Nearest Neighbours (kNN), Random Forest, and Artificial Neural Networks (ANN), have been widely used in previous research to interpret PPG signals~\cite{r5, r6, r7, r8, r9}. However, the process of data labeling is tedious, time-consuming, and costly, especially for anomaly detection problems, since anomalies seldomly occur in real-world applications. Furthermore, these supervised learning methods may be prone to bias and overfitting if the labeled dataset does not adequately represent the full range of normal and anomalous PPG signals. Moreover, these methods may not be adaptable to unknown or unexpected anomalies, as they learn to recognize patterns based on the examples provided in the training dataset, and may fail to effectively detect anomalies not represented in the dataset. To address these limitations, unsupervised anomaly detection methods can offer advantages over supervised approaches, as they do not rely on explicitly labeled examples of anomalous behavior and can be more adaptable to unknown or unexpected anomalies~\cite{r9_added_1, r9_added_2}.

Anomaly detection in PPG data can also be challenging due to other various factors that contribute to noise and inter-subject variability. These are factors like physical activity, stress, illness, measurement noise, age, gender, body composition, and genetic differences, as well as external factors such as sensor placement, sensor quality, and environmental conditions. This makes it difficult to develop generalized models that perform consistently across different individuals since each person's PPG signal may exhibit unique characteristics~\cite{r4_added_1}. These complexities necessitate strategies to account for individual-specific characteristics.

Personalization can be a potential solution to help overcome the limitations of generalization by tailoring models to individual users~\cite{r4_added_2}. However, the effectiveness of personalization hinges on accurate biometric identification. Inaccurate identification of individuals can lead to personalized models being trained on or applied to the wrong user's data, resulting in poor performance and potentially harmful outcomes. Hence, accurate biometric identification can enhance the reliability of personalized models, as it ensures that the models are based on the specific characteristics of each user.

In addition to unsupervised anomaly detection methods and personalization, representation learning can be particularly useful in enhancing performance~\cite{r9_added_3}. Representation learning models are typically trained to learn from large amounts of unlabeled data, enabling them to extract a more compact, informative, and expressive representation of the data without the need for expensive and time-consuming manual labeling. By learning a lower-dimensional representation of the PPG data that captures its inherent structure and discriminative features, representation learning can help overcome challenges posed by inter-subject variability, noise, and other factors affecting PPG signals. AutoEncoders, for example, are a type of self-supervised representation learning model that learns representations by encoding inputs into lower dimensions and then decoding them back to their original form, focusing on reconstructing the input~\cite{r10}. Other representation learning models have been proposed with different tasks, such as contrastive learning or classification of augmented transformations of the original data~\cite{r11,r12}. Representation learning has been increasingly used for anomaly detection in various domains, including image analysis~\cite{r13, r14, r15, r16, r17}, and time series data, such as bio-signals sensor data like EEG or ECG~\cite{r18, r19}. However, its application to PPG data for anomaly detection and biometric identification remains underexplored, despite PPG being a commonly used bio-signal in health-monitoring applications.\\

\noindent  In this paper, we present a two-stage framework for unsupervised movement detection and biometric identification in PPG data using representation learning. In the first stage, we train a deep neural network to obtain a lower-dimensional and informative data representation. In the second stage, we construct separate unsupervised anomaly detectors for both tasks using the learned representations from the first stage. Our approach not only investigates the effectiveness of representation learning in this context, but also explores the potential of personalization in enhancing anomaly detection performance. Additionally, we delve into biometric identification, aiming to improve the reliability of personalized anomaly detectors. To the best of our knowledge, this is the first study to jointly address these aspects for anomaly detection in PPG data. Summarizing, our contributions are:

\begin{enumerate}
\item We propose a two-stage framework for unsupervised anomaly detection and biometric identification in PPG data using representation learning.
\item We demonstrate the effectiveness of using the learned representations compared to the original representations in detecting difficult real-world anomalies.
\item We compare the effectiveness of generalization and personalization in anomaly detection, discussing the impact of tailoring models to individual users.
\item We investigate the unsupervised biometric identification task in PPG data to increase the reliability of personalized models.
\item We explore the impact of the dimensionality of the learned representation on the performance of our anomaly detection framework, demonstrating the robustness of representation learning across a wide range of dimensionalities.
\end{enumerate}

\section{Proposed Framework}
\begin{figure}[th!]
    \centering
    \includegraphics[width=\textwidth]{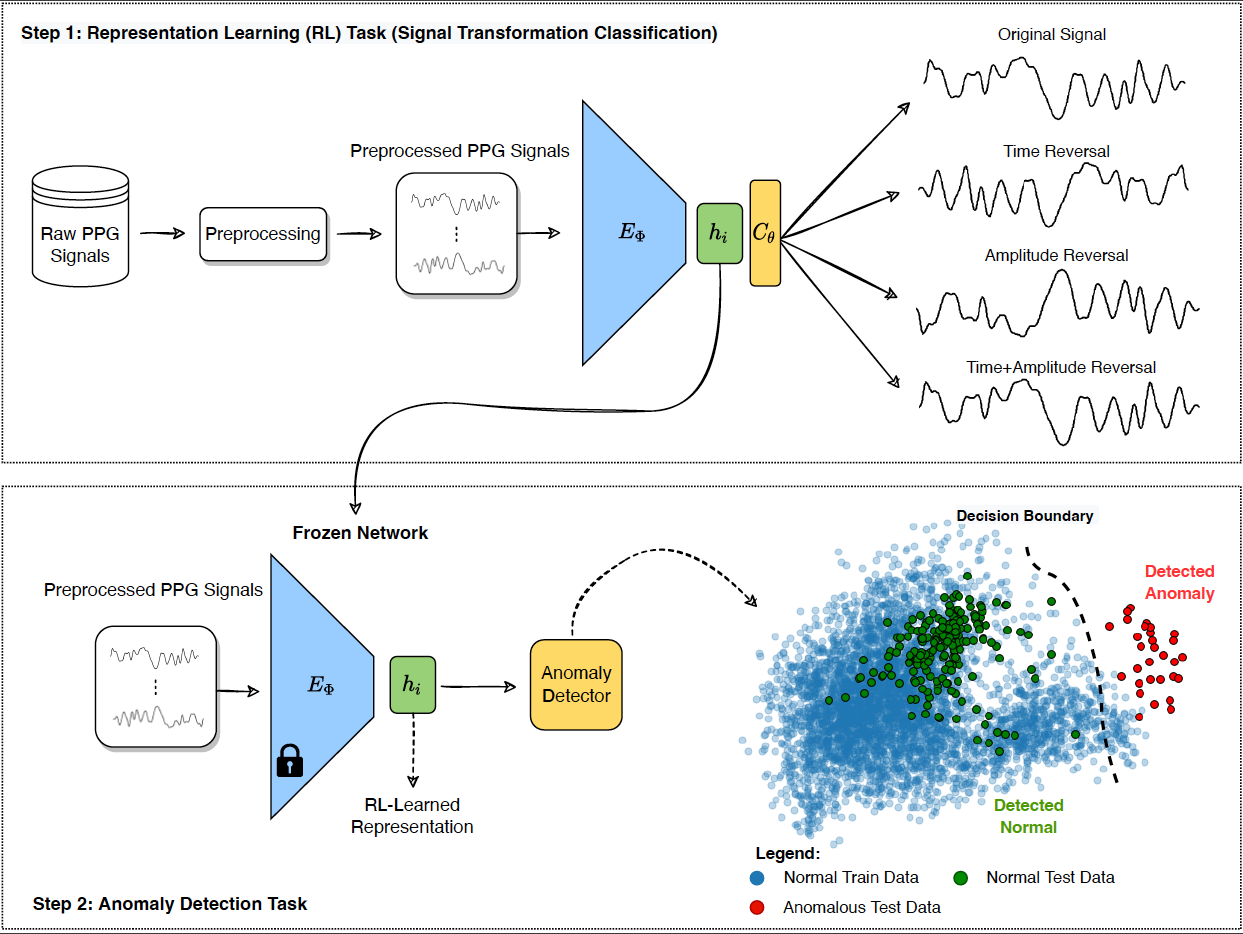}
    \caption{Proposed framework for anomaly detection using Representation Learning (RL). The framework consists of two steps: (1) A representation learning phase, where the model (consisting of an encoder and classifier component) is trained to discriminate between augmented transformations and the original data. The weights of the encoder are then frozen for the next step. (2) A anomaly detection phase, where the frozen encoder is used to extract features from the input data, which are then fed into an anomaly detector. 
    The scatter plot illustrates an example of the distribution of points in a 2-dimensional feature space. The anomaly detector separates the normal and anomalous samples with the decision boundary (threshold) based on their anomaly scores. }
    \label{Proposed_Framework}
\end{figure}

An overview of the proposed anomaly detection framework is shown in Figure~\ref{Proposed_Framework}. In the first step, we focus on obtaining a representation of the PPG data that captures the underlying structure of the data. Recent research shows that the task of classifying the original data and augmented transformed versions of the same data can outperform AutoEncoders and contrastive learning methods in learning better representation for the downstream task of interest~\cite{r12}. Accordingly, we learn the representation by distinguishing original data from augmented transformed versions of the same data. This task is what we refer to as "Signal Transformation Classification."

Given the original signal $S(l)$, where $l = (1, 2,\ldots, L)$ and $L$ is the length of the time series, the augmented transformations of the data are described as:

\begin{itemize}[itemindent=1.5em]\itemsep0.5em
    \item \emph{Time reversal}: A time inverted version of the signal: $S'(l)$, where $l = (L, L-1,\ldots,1)$.
    
    \item \emph{Amplitude reversal}: A amplitude inverted version of the signal: as $S'(l)=-S(l)$, where $l = (1,2,\ldots,L)$.

    \item \emph{Both Time and Amplitude reversal}: We first perform the time reversal as described and then perform the amplitude reversal to obtain a time and amplitude inverted version of the signal: $S'(l)=-S(l)$, where $l = (L, L-1,\ldots,1)$.
    
\end{itemize}

\noindent To train the representations, we use a CNN model to classify PPG segments into four categories: Time reversal, Amplitude reversal, Both time and amplitude reversal, and the original signal. Given an unlabeled PPG dataset $D_U=\left\{ \textbf{x}_{i} \right\}_{i=1}^{N_u}$ where $\textbf{x}_{i}\in \mathbb{R}^{1\times T}$ is a vector of length $T$ and $N_u$ is the number of vectors (samples). $y_i \in \{1, 2, 3, 4\}$ is the class label for the $i^{th}$ vector, where $y_i=1, 2, 3$ represents the augmented data obtained by reversing the original PPG signal and $y_i=4$ represents the original PPG signal. The CNN model consists of an encoder component that maps each input vector into a latent space representation $\textbf{h}_i=E_\phi(\textbf{x}_i)$ where $\textbf{h}_{i}\in \mathbb{R}^{1\times d}$ and $d< T$. After that, $\textbf{h}_i$ is fed into the classifier component of the model to predict the class label $\hat{y}_i=C_\theta (\textbf{h}_i)$. The model is trained to minimize the cross-entropy loss between the predicted class label $\hat{y}_i$ and the true class label $y_i$. The final learned representation is obtained by taking the latent space representation $\textbf{h}_i$ outputted by the encoder component. This learned representation is then used in the second stage of our proposed framework for anomaly detection.\\

\noindent In the second step of our proposed framework, we use the learned representation $\textbf{h}_{i}$ to detect anomalies. Specifically, we use three different methods to detect whether an input signal is an anomaly: Multi-Variate Normal distribution (MVN)~\cite{r19_added_1}, Isolation Forest (IF)~\cite{r19_added_2}, and PCA-Reconstruction~\cite{r19_added_1}. For the MVN, the mean and covariance matrix are estimated on normal training samples. Given a test sample $\textbf{h}_{test}$, we can then calculate the probability density function (PDF) of the test sample using the fitted Gaussian distribution as:

\begin{equation}
p(\textbf{h}_{test}) = \frac{1}{(2\boldsymbol{\pi})^{d/2}|\boldsymbol{\Sigma}|^{1/2}} \exp\left( -\frac{1}{2}(\textbf{h}_{test} - \boldsymbol{\mu})^T \boldsymbol{\Sigma}^{-1} (\textbf{h}_{test} - \boldsymbol{\mu}) \right)
\end{equation}

\noindent where $d$ is the dimension of the learned representation. It is expected that anomalous test samples have a lower probability compared to normal samples. Therefore, these points can be detected if the probability is below a set threshold. 

For the Isolation Forest (IF) method, we first train an ensemble of decision trees on normal training samples. Given a test sample $\textbf{h}_{test}$, the IF algorithm isolates the test sample from the others by recursively splitting the data with randomly selected features and split values. The number of splits, or the path length, required to isolate a sample is an indication of its anomaly score. Anomalous samples are expected to have shorter path lengths compared to normal samples. The anomaly score of a test sample $\textbf{h}_{test}$ using the IF algorithm is calculated as:

\begin{equation}
s(\textbf{h}_{test}) = 2^{-\frac{\textstyle E[L(\textbf{h}_{test})]}{\textstyle c(N)}}
\end{equation}

\noindent where $E[L(\textbf{h}_{test})]$ is the average path length of the test sample over all trees in the ensemble, $c(N)$ is the average path length of an unsuccessful search in a Binary Search Tree with $N$ external nodes, and $N$ is the number of samples in the training data. The anomaly score $s(\textbf{h}_{test})$ ranges from 0 to 1, with higher scores indicating a higher likelihood of being anomalous. Anomalous test samples can be detected if the anomaly score is above a set threshold. \\

\noindent The PCA-Reconstruction method is a technique for detecting anomalies in high-dimensional data by reconstructing the original data from its principal components and evaluating the reconstruction error. Given a test sample $\textbf{h}_{test}$, the reconstruction error can be calculated as the squared distance between the original sample and its reconstructed version ($\textbf{h}_{recon}$) after mapping to a reduced PCA space. This is achieved by projecting the test sample $\textbf{h}_{test}$ onto the orthogonal basis vectors represented by the matrix describing the PCA mapping, $W$, and then transforming it back to the original space. The reconstruction error can then be expressed as: 
\begin{equation}
e(\textbf{h}_{test}) = ||\textbf{h}_{test} - (WW^T)\textbf{h}_{recon}||^2
\end{equation}

\noindent Note that anomalous test samples can be detected if the reconstruction error is above a set threshold. 

\subsection{Definition of Anomalies}
We define anomalies in the context of two specific tasks: activity movement detection and biometric identification. 

\subsubsection{Activity Movement Detection}
In this particular setting, we train an anomaly detector on the recorded data during a specific activity (considered as the "normal" activity) and evaluate it on the data, which includes another activity (considered as an "anomalous" activity) in addition to the "normal" activity. We assume that the anomalous movement activity shows a different pattern than the normal activity and should be distinguishable from the "normal" movement activity. Accurately detecting movement can have significant practical implications in various applications, such as fitness health tracking, where identifying irregular patterns or deviations from expected behavior is crucial. By focusing on such a complex and practical problem, we can demonstrate the effectiveness and robustness of our proposed approach in handling real-world challenges associated with PPG data, including inter-subject variability, noise, and other factors affecting signal quality.

\subsubsection{Biometric Identification}
In the context of biometric identification, we aim to identify an individual (user) as an anomaly when compared to a given group of people or another individual as the "intended" user(s). We train the anomaly detector on the recorded data from the intended group or individual during a specific activity and evaluate the anomaly detector when presenting new data, which includes another individual (considered as an "anomaly") during the same activity as the data from the intended user(s). Identifying such anomalies can be crucial in personalized health monitoring systems, where it is important to distinguish between users for accurate and safe health monitoring and assessments.

\section{Experimental Setup}
\subsection{Datasets}
We use two datasets in our experiments. The first one is the Pulse Transit Time PPG (PTT-PPG) public dataset~\cite{r20}, a high-resolution and time-synchronized dataset annotated with activity labels. It contains waveform records from multi-wavelength sensors measuring PPGs, attachment pressures, and temperatures. The recordings are from 22 healthy subjects (M = 22) performing different physical activities in random order. We selected \emph{Sitting} and \emph{Walking} activities for this study. We use the green wavelength recorded PPG from the proximal phalanx (base segment) of the left index finger palmar side (Frequency of 500~Hz). 

The second dataset is the PPG-Dalia public dataset collected by \citet{r21} to perform PPG-based heart rate estimation. It has recordings of 15 subjects (M = 15) performing different daily activities. We have selected \emph{Sitting} and \emph{Walking} activities for this study. We removed data from subject number 6 due to incomplete data recording. The signals are recorded with a frequency of 64~Hz.

\subsection{Data Preprocessing }
A band-pass $2^{nd} $ order Butterworth filter is applied to the whole PPG signal of each subject individually for both datasets, but with different frequency ranges of 0.35 - 20~Hz for the PTT-PPG dataset and 0.1-10~Hz for PPG-Dalia. To create different categories of signals, we used Time reversal, Amplitude reversal, and both Time and Amplitude reversal augmentations. All of the signals are then normalized to zero mean and unit variance across the whole signal per subject. The final normalized filtered signals are segmented into windows with a length of 8 seconds, while two successive windows overlap by 7.5 seconds (this setting is common for PPG data~\cite{r21,r22, r23}). Since the PTT-PPG dataset frequency is 500~Hz, the input windows are resampled using the Fourier method from a size of 4000 to a fixed size of 512, which allows for more efficient processing during model training, and it is the same input window size as the PPG-Dalia dataset.

\subsection{Implementation} 
\subsubsection{Representation Learning}
The hyperparameters and the architecture of the proposed deep learning model are determined by systematically searching through all possible combinations to obtain the best performance on the classification task using Leave-One-Subject-Out cross-validation (LOSO). Eventually, we used a CNN architecture deep learning model consisting of a 1D convolutional neural network layer with a series of five-layer blocks followed by a fully connected layer and a final classification layer. The layer blocks are composed of two 1D convolutional layers, each followed by the Exponential Linear Unit (ELU) activation function and, in the end, a MaxPooling layer. After the final layer block, there is a fully connected layer with a size of 64, which is the learned representation size, followed by the Rectified Linear Unit (ReLU) activation function. Finally, there is a classification layer (SoftMax activation function) with a size of 4, corresponding to the four categories. The final implemented CNN model details are available in Appendix A.

The model is optimized using categorical cross-entropy as the loss function. The Adam optimizer is used with a learning rate of 0.00001 and a decay rate of 0.0001 for the PTT-PGG dataset and a learning rate of 0.0001 and a decay rate of 0.001 for the PPG-Dalia dataset. The batch size is 64, and training runs for 400 epochs for both datasets. To assess the randomness of the deep learning framework, each training process for each test subject is repeated five times. To evaluate the signal transformation classification performance, we use the Area under the ROC curve (AUC-ROC) metric.

\subsubsection{Anomaly Detection}
In our PCA-based anomaly detection approach, we optimize the number of principal components by ensuring they cumulatively account for 99\% of the data variance. The Isolation Forest model was implemented with 100 base estimators in the ensemble. The number of base estimators was chosen based on our preliminary experiments, which showed good performance in this setting. The Multivariate Normal Distribution-based anomaly detector was implemented utilizing a Gaussian Mixture Model with a single component. The parameters of this distribution, namely the mean vector and the covariance matrix, are learned directly from the data. In the evaluation phase, we assess the performance of our anomaly detectors by calculating the AUC-ROC.

\subsection{Anomaly Detection Evaluation Scenarios}
We consider two evaluation scenarios for anomaly detection tasks: Generalization and Personalization. 

\subsubsection{Generalization Scenario}
In the generalization scenario, we aim to test the ability of the anomaly detection model to generalize across different individuals.

For activity movement detection, shown in Fig~\ref{fig2_generalization}.a, we train the model using data from all subjects performing \emph{Sitting} activity as the main normal activity. This data is considered as 'normal training samples'. In the test phase, we introduce data from both a new activity, referred to as the 'anomalous activity' (in this case, \emph{Walking}), and the main activity of a new subject (left-out) who was not part of the training data. We repeat this process for each subject, treating them as the test set (left-out subject), using the LOSO setting. We then calculate the mean and standard deviation of the performance metrics across all test sets.

For biometric identification, shown in Fig~\ref{fig2_generalization}.b, we train the model on data from a group of subjects, who we refer to as the 'intended users'. This data forms our 'normal training samples'. We set aside 20\% of the data from each subject for testing, using a 5-fold cross-validation approach. During the testing phase, we introduce 'anomalous data' from a new subject who is not part of the training data. This subject is referred to as the 'left-out' subject (user). To assess how well our model can differentiate the new user from the intended users, we use LOSO validation to treat each subject once as a 'left-out' user. We then calculate the mean and standard deviation of the performance metrics across all test sets.

\input{Figures/Fig2_Generalization}

\input{Figures/Fig3_Personalization}

\subsubsection{Personalization Scenario}
In the personalization scenario, we aim to tailor the anomaly detection model to individual characteristics, both for movement detection and biometric identification.

\noindent For activity movement detection, shown in Fig~\ref{fig3_personalization}.a, we select one subject and train the model on data related to the main activity, which is \emph{Sitting}. This data forms our 'normal training samples'. We reserve 20\% of this data for testing, using a 5-fold cross-validation approach. During the testing phase, we introduce 'anomalous activity data' from the same subject, in this case, \emph{Walking} activity, and we use 20\% of this data using 5-fold cross-validation for testing. Thereby, our test set includes 20\% of '\emph{Walking} activity' data and 20\% of the '\emph{Sitting} activity' data from one selected subject. We repeat this process for each subject. Finally, we calculate the mean performance and standard deviation across all test sets.

For biometric identification, shown in Fig~\ref{fig3_personalization}.b, we train the model using data from a single selected subject, who we refer to as the 'intended user'. This data forms our 'normal training samples'. We reserve 20\% of this data for testing, using a 5-fold cross-validation method. During the testing phase, we introduce 'anomalous data' from a new subject who was not part of the training data, and we use 20\% of its data using 5-fold cross-validation for testing.
We compare each subject with the intended user in a pairwise manner. The average performance of these comparisons is taken as the performance of the intended user. We repeat this entire process for each individual, treating them as the intended user each time. Finally, we calculate the mean performance and standard deviation across all intended individuals.

\section{Results}
\subsection{Representation Learning}
The first step of the proposed framework is Representation Learning (see Fig~\ref{Proposed_Framework}). The overall performance of the signal transformation classification task for both datasets is calculated across all test subjects. Both datasets have a high mean AUC of $0.92 \pm 0.09$ for the PTT-PPG and $0.93 \pm 0.06$ for the PPG-Dalia. These results indicate that the model is able to generalize to new data (subject) and accurately classifies the augmented and original PPG segments in both datasets.  

\subsection{Anomaly Detection}

\subsubsection{Activity Movement Detection}
The second step of the proposed framework is Anomaly Detection (see Fig~\ref{Proposed_Framework}). Table~\ref{Table_Complete_AD} shows the results of movement detection for both datasets. In the generalized scenario, representation learning significantly improves the AUC performance for all three anomaly detection methods, suggesting its effectiveness in detecting anomalies in PPG data compared to the original data representation. For instance, the results of all anomaly detectors with PTT-PPG reveal that the AUC performance for anomaly detection barely reaches 0.5. However, the performance is increased towards 0.9 when using the learned representations. 

Further, one can see that the models are typically very unstable in the original representation, suggesting that there is high inter-subject variability. The results demonstrate that representation learning helps decrease this variability, enabling better generalization across different individuals. \\

\noindent We also investigated the performance of the proposed methods in a personalized setting (Table~\ref{Table_Complete_AD}). The performance of all methods in the personalized setting is significantly higher than the generalized situation, with reduced variability among individual performances. This indicates that personalization improves performance while maintaining consistency across different individuals. Representation learning continues to outperform the original data representation in the personalized setting, emphasizing the effectiveness of learned representations in capturing subject-specific characteristics of PPG data. 

\input{Tables/Table_Complete_AD}

\subsubsection{Biometric Identification}
In light of the improved performance achieved through personalization in activity movement detection, biometric identification ensures that the detected anomalies are specific to the intended user. Table~\ref{Table_Complete_BI} shows the results of biometric identification in both generalized and personalized scenarios during \emph{Sitting} Activity. In the generalized scenario, it can be observed that using representation learning is effective, and it improves the performance of all anomaly detectors across both datasets. While representation learning has been successful in improving performance, it still may not be perfect. This can be attributed to the inter-subject variability present in the data, as the model must distinguish between multiple people considered normal, which is challenging.

\noindent Considering the personalized scenario results, the performance of all methods is significantly higher compared to the generalized scenario, with substantially reduced variability: while in the generalized scenario, the standard deviations of the results are often around 0.2, in the personalized scenario, it is reduced to below 0.1. Moreover, representation learning continues to improve performance in the personalized setting, demonstrating the effectiveness of learned representations. These results indicate that minimizing inter-subject variability allows the model to better identify the anomalous person as it is easier to detect the anomalous individual from only one individual compared to a group.

\input{Tables/Table_Complete_BI}

\subsection{Robustness of Representation Dimensionality}

One key aspect of our anomaly detection framework is the dimensionality of the learned representation, denoted as $\textbf{h}_{i}$. Figure~\ref{fig5_RepAnalysis} illustrates the mean performance of anomaly detectors with varying $\textbf{h}_{i}$ dimensions ranging from 2 to 512 for both datasets in generalized and personalized scenarios.

As the $\textbf{h}_{i}$ dimensionality increases, the AUC also increases up to a certain point. This trend suggests that as the dimensionality rises, the representation captures more valuable information for anomaly detection. However, once we reach a certain dimensionality, further increases do not provide additional benefits, and the performance is stable. 

In both scenarios, the learned representation improves the AUC compared to the original signal.  Results show that using learned representation leads to better performance when the dimensionality of $\textbf{h}_{i}$ is reduced to extremely low levels. For instance, at the low dimensionality of 2 in PTT-PPG and 8 in the Dalia datasets, we can see the improvements in using learned representation over the original signal. Even when the dimensionality of the learned representation is the same as the original signal's dimensionality (512), it outperforms the original signal. This robust performance of the representation learning approach highlights its effectiveness in capturing the essential structure and patterns of the data and learning useful features across a wide range of low and high dimensionalities. 

Choosing the right dimensionality depends on a balance between model performance and computational efficiency. Based on the results, the dimensionality of 64 for the PTT-PPG and 256 for the Dalia dataset seems to offer an ideal balance between computational efficiency and performance.

\input{Figures/Fig5_RepAnalysis}

\section{Discussion and Conclusion}
This paper proposes a framework for anomaly detection in PPG data, consisting of two stages: representation learning and anomaly detection (Activity Movement Detection and Biometric Identification). We tested the ability of the proposed framework in generalized and personalized scenarios. The results showed that representation learning and personalization both are effective approaches for improving the performance of anomaly detection, potentially allowing for the detection of rare and subtle heart conditions. 

The activity movement detection results emphasize the effectiveness of representation learning by improving the AUC performance and decreasing the inter-subject variability. Nevertheless, in some cases (Dalia dataset), the variability was not significantly decreased. This could be attributed to a variety of factors, including the inherent complexities of the dataset and the nature of the anomaly detection methods themselves. The inherent characteristics of the dataset, such as signal quality, the sensor used for data collection, the environmental conditions during data collection, and the demographic characteristics of the participants, can impact representation learning. These factors can make it more challenging for the model to learn robust representations, thereby leading to higher variability. Note that the original representation AUC results, which are lower than 0.5, may indicate that flipping the label is actually beneficial. For example, it can be seen in Table\ref{Table_Complete_AD} that an AUC of 0.28 in the generalized scenario from the PTT-PPG dataset would significantly improve to 0.72 by flipping, but it still remains worse than the 0.91 obtained by the RL representation. These results suggest that our approach may be beneficial in real-world applications.

Although the performance improved by representation learning, the challenge of decreasing the variability in all datasets suggests the potential for further improvements through personalization. Personalization emerged as an essential factor in improving the performance of all methods in detecting activity movements. By tailoring models to individual users, the system was better able to capture subject-specific characteristics, leading to reduced variability among individual performances. These findings underscore the importance of personalization in building reliable and accurate PPG-based health monitoring systems.

Biometric identification plays a crucial role in ensuring that personalized models are specific to the intended user. While representation learning improved the performance of all anomaly detectors in both generalized and personalized scenarios, the performance in the generalized scenario was not perfect. This limitation can be attributed to the inter-subject variability present in the data, which poses challenges for distinguishing anomalous subject among a group of subjects considered normal. In contrast, the personalized scenario demonstrated significantly higher performance and substantially reduced variability, as the model could better identify the anomalous subject when focusing on a single individual.

Analyzing the robustness of representation dimensionality underscores its significance in anomaly detection frameworks. The performance of anomaly detection in relation to the dimensionality of the learned representation follows a pattern of initial gains followed by a plateau. This pattern suggests that while increasing dimensionality can enhance performance, there is a threshold beyond which additional increases do not yield further benefits. Interestingly, at the same dimensionality as the original signal, representation learning performs better. This suggests that learned representations can capture the (nonlinear) underlying patterns or structures in the data that may not be immediately apparent in the original signal. Furthermore, the fact that the learned representation can outperform the original signal even at extremely low dimensionalities signifies that representation learning can effectively extract and retain the most critical information from the original signal, thereby enhancing anomaly detection.

In all experiments of our framework, the performance difference between anomaly detection methods was relatively small. Therefore, we cannot draw a clear conclusion about which method performs better than the others overall. It seems that the crucial point in anomaly detection is not the method, but it is the representation and the personalization. \\

\noindent Despite the promising results in using representation learning and personalization, it is important to note that further research is needed to evaluate the effectiveness of RL on a wider range of different types of real-world anomalies in PPG. This is particularly important for practical applications, such as using smartwatches and self-monitoring for anomaly detection in healthcare, where the complexity and variability of real-world anomalies may be high. Exploring different algorithms or techniques to enhance the learned representations of PPG data can also be a future direction to further improve anomaly detection performance and decrease inter-subject variability.

In conclusion, our proposed framework provides a promising approach for different types of anomaly detection in PPG data. Combination of representation learning and personalization provides a more effective approach for developing reliable, robust, and accurate health monitoring systems.\\

\section*{Acknowledgment}
Funding: This work was supported by the Dutch Research Council (NWO) [grant numbers 628.011.214].

\bibliographystyle{unsrtnat}
\renewcommand{\bibname}{\protect\leftline{References}}
\renewcommand{\bibfont}{\small}
\bibliography{ref} 

\newpage
\appendix

\section{Deep Learning Model Architecture for Representation Learning Task}
\begin{figure}[th!]
    \centering
    \renewcommand{\thefigure}{A.1} 

    \includegraphics[width=\textwidth]{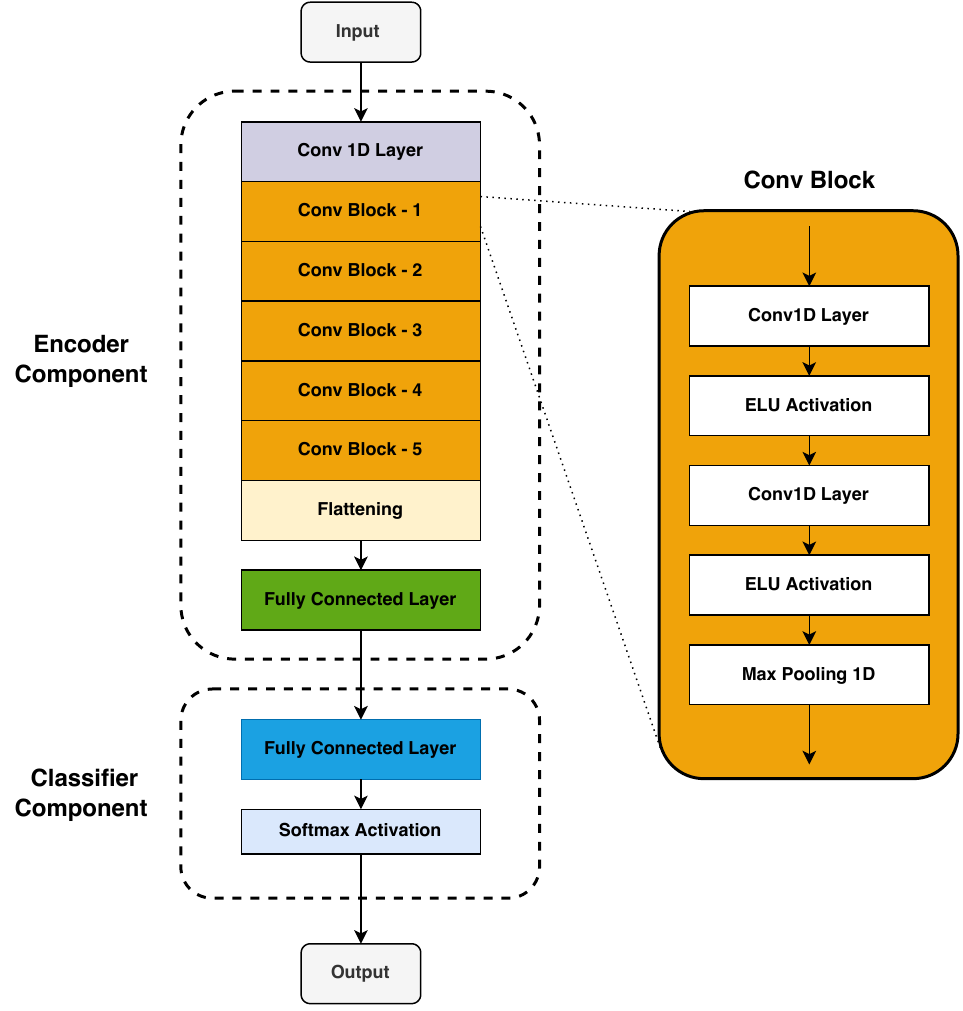}
    \caption{Outline of the Implemented Encoder-Classifier Deep Learning Model}
    \label{Model_Visualization}
\end{figure}

The Deep Learning framework is developed and evaluated in Python (Version 3.8.8) using Keras API. It should be noted that for the parameters that are not mentioned in the implementation details, the Keras default settings are used. 
The detailed architecture and outline of the implemented model for the representation learning task is shown in Table~\ref{tab:tableA} and Figure~\ref{Model_Visualization}.

\input{Tables/TableA.tex}

\end{document}

%% file: Figures/Fig2_Generalization.tex
\begin{figure}[ht]
    \centering
    \subfigure[\centering]{{\includegraphics[width=7cm, height=5cm]{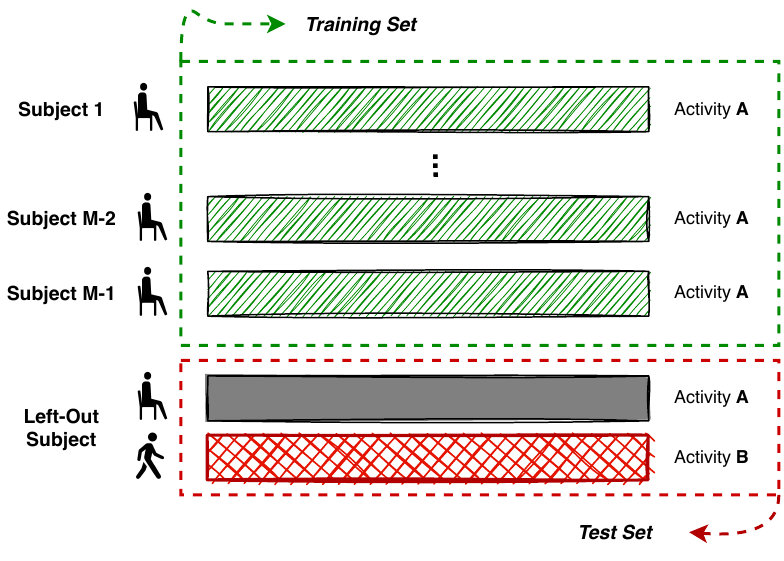} }}%
    \qquad
    \hspace{1mm}
    \subfigure[\centering]{{\includegraphics[width=7.7cm, height=5cm]{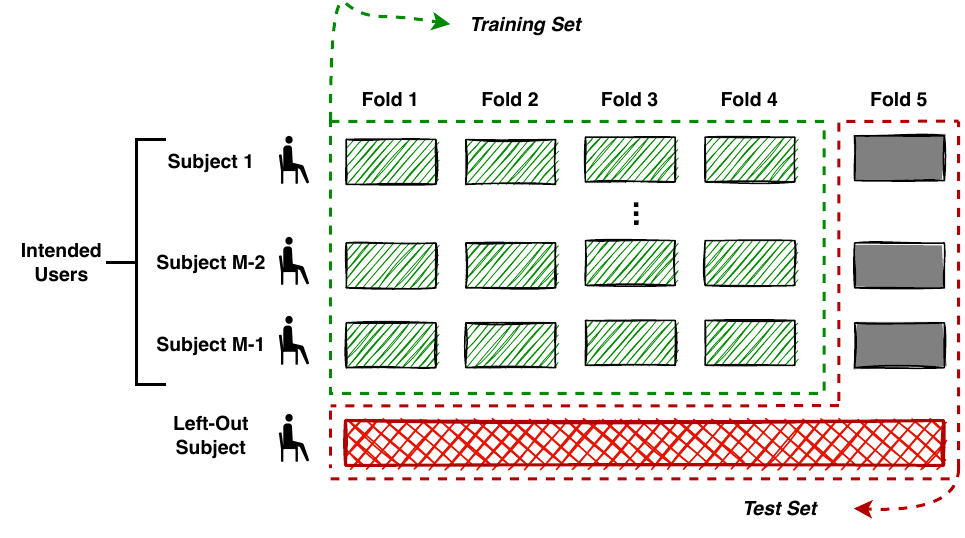} }}%

    \begin{minipage}{\textwidth}
    \centering
    \small
    
    \tikz{\draw [OliveGreen, pattern=north east lines, pattern color=OliveGreen] (0,0) rectangle (0.6cm,0.3cm);}\hspace{2mm}Training samples (Normal), \hspace{2mm} 
   \tikz{\draw[fill=gray, draw=black] (0,0) rectangle (0.6cm,0.3cm);}\hspace{2mm}Test samples (Normal), \hspace{2mm}   \tikz{\draw [red, pattern=crosshatch, pattern color=red] (0,0) rectangle (0.6cm,0.3cm);}\hspace{2mm}Test samples (Anomaly)
    
    \end{minipage}
    
    \caption{Overview of Generalization Scenario a) Generalization in Movement Detection task b) Generalization in Biometric Identification task. Note that the distribution of the anomalous and normal samples in training and test sets follows the same ratios as depicted in the figures. 
}%
    \label{fig2_generalization}
\end{figure}

%% file: Figures/Fig3_Personalization.tex
\begin{figure}[h!]
    \centering
    \subfigure[\centering]{{\includegraphics[width=7cm, height=5cm]{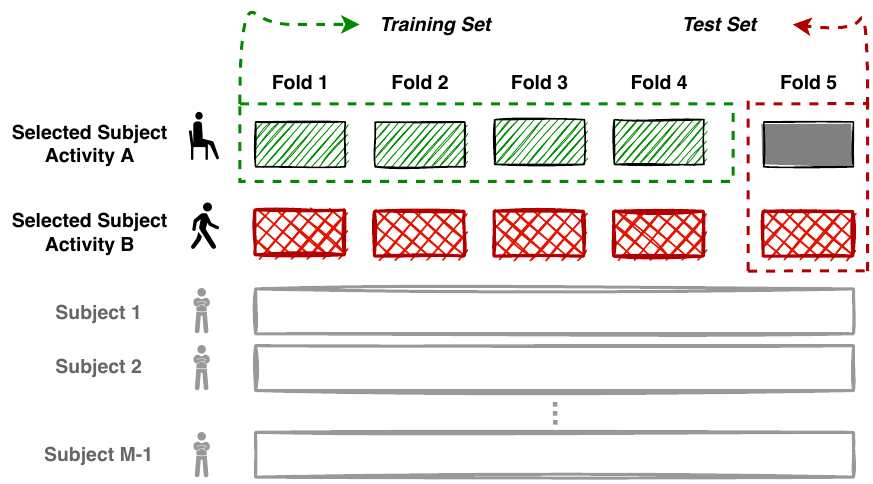} }}%
    \qquad
    \hspace{5.5mm}
    \subfigure[\centering]{{\includegraphics[width=7cm, height=5cm]{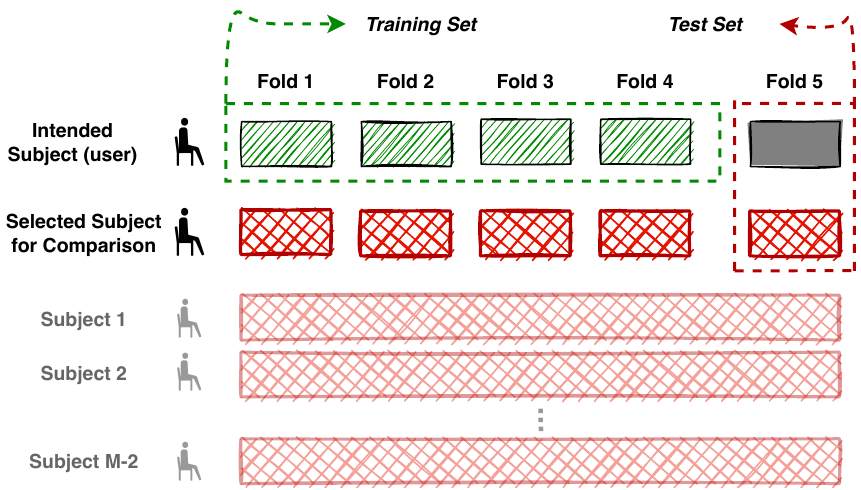} }}%

    \begin{minipage}{\textwidth}
    \centering
    \small
    
    \tikz{\draw [OliveGreen, pattern=north east lines, pattern color=OliveGreen] (0,0) rectangle (0.6cm,0.3cm);}\hspace{2mm}Training samples (Normal), \hspace{2mm} 
   \tikz{\draw[fill=gray, draw=black] (0,0) rectangle (0.6cm,0.3cm);}\hspace{2mm}Test samples (Normal), \hspace{2mm}   \tikz{\draw [red, pattern=crosshatch, pattern color=red] (0,0) rectangle (0.6cm,0.3cm);}\hspace{2mm}Test samples (Anomaly)
    
    \end{minipage}

    \caption{Overview of Personalization Scenario a) Personalization in Movement Detection task. b) Personalization in Biometric Identification task. Note that the distribution of the anomalous and normal samples in training and test sets follows the same ratios as depicted in the figures.}%
    \label{fig3_personalization}
\end{figure}

%% file: Tables/Table_Complete_AD.tex
\begin{table}[h!]
\centering
\caption{Mean Test AUC performance of Movement Detection in the Generalized and Personalized Scenario. The 'RL-' prefix designates anomaly detectors that employ learned representations from Representation Learning (RL) instead of the original data representation. }
\label{Table_Complete_AD}

\begin{tabular}{p{3.5cm}P{3cm}P{3cm}P{3cm}P{3cm}} 
\toprule
\multirow{3}{3cm}{Anomaly Detectors} & \multicolumn{4}{c}{\textbf{Movement Detection AUC Performance}} \\ \cline{2-5}
                                & \multicolumn{2}{c}{\textbf{PTT-PPG Dataset}} & \multicolumn{2}{c}{\textbf{Dalia Dataset}} \\ \cmidrule{2-3} \cmidrule{4-5}
                                & \textbf{Generalized} & \textbf{Personalized} & \textbf{Generalized} & \textbf{Personalized} \\ \midrule

MVN        & $0.40 \pm 0.39$      & $0.74 \pm 0.29$    & $0.78 \pm 0.16$   & $0.93 \pm 0.08$    \\
RL-MVN    & \textbf{0.92 $\pm$ 0.09} &   \textbf{0.98 $\pm$ 0.03} & \textbf{0.93 $\pm$ 0.10} & \textbf{0.97 $\pm$ 0.03} \\[2mm]

IF       & $0.28 \pm 0.34$      & $0.37 \pm 0.31$    & $0.56 \pm 0.14$   & $0.87 \pm 0.15$    \\
RL-IF    & \textbf{0.91 $\pm$ 0.11} &   \textbf{0.97 $\pm$ 0.05} & \textbf{0.88 $\pm$ 0.13} & \textbf{0.93 $\pm$ 0.03} \\[2mm]

PCA      & $0.44 \pm 0.37$      & $0.81 \pm 0.25$    & $0.76 \pm 0.19$   & $0.94 \pm 0.07$    \\
RL-PCA   & \textbf{0.91 $\pm$ 0.09} &   \textbf{0.97 $\pm$ 0.03} & \textbf{0.90 $\pm$ 0.18} & \textbf{0.95 $\pm$ 0.03} \\[2mm]

\bottomrule
\end{tabular}
\end{table}

%% file: Tables/Table_Complete_BI.tex

\begin{table}[h!]
\centering
\caption{Mean Test AUC performance of Biometric Identification in the Generalized and Personalized Scenarios. Results are based on \emph{Sitting} Activity. The 'RL-' prefix designates anomaly detectors that employ learned representations from Representation Learning (RL) instead of the original data representation.  }
\label{Table_Complete_BI}

\begin{tabular}{p{3.5cm}P{3cm}P{3cm}P{3cm}P{3cm}} 
\toprule
\multirow{3}{3cm}{Anomaly Detectors} & \multicolumn{4}{c}{\textbf{Biometric Identification AUC Performance}} \\ \cline{2-5}
                                & \multicolumn{2}{c}{\textbf{PTT-PPG Dataset}} & \multicolumn{2}{c}{\textbf{Dalia Dataset}} \\ \cmidrule{2-3} \cmidrule{4-5}
                                & \textbf{Generalized} & \textbf{Personalized} & \textbf{Generalized} & \textbf{Personalized} \\ \midrule

MVN        & $0.40 \pm 0.26$      & $0.76 \pm 0.22$    & $0.43 \pm 0.26$   & $0.56 \pm 0.20$    \\
RL-MVN    & \textbf{0.60 $\pm$ 0.22} &   \textbf{0.86 $\pm$ 0.08} & \textbf{0.55 $\pm$ 0.17} & \textbf{0.78 $\pm$ 0.09} \\[2mm]

IF       & $0.45 \pm 0.36$      & $0.58 \pm 0.29$    & $0.45 \pm 0.29$   & $0.56 \pm 0.24$    \\
RL-IF    & \textbf{0.61 $\pm$ 0.24} &   \textbf{0.86 $\pm$ 0.08} & \textbf{0.53 $\pm$ 0.18} & \textbf{0.74 $\pm$ 0.09} \\[2mm]

PCA      & $0.39 \pm 0.34$      & $0.67 \pm 0.24$    & $0.43 \pm 0.26$   & $0.55 \pm 0.22$    \\
RL-PCA   & \textbf{0.59 $\pm$ 0.20} &   \textbf{0.84 $\pm$ 0.09} & \textbf{0.55 $\pm$ 0.16} & \textbf{0.78 $\pm$ 0.09} \\[2mm]

\bottomrule
\end{tabular}
\end{table}

%% file: Figures/Fig5_RepAnalysis.tex
\begin{figure}[ht]
    \centering
    \subfigure[\centering]{{\includegraphics[width=7.3cm, height=5.3cm]{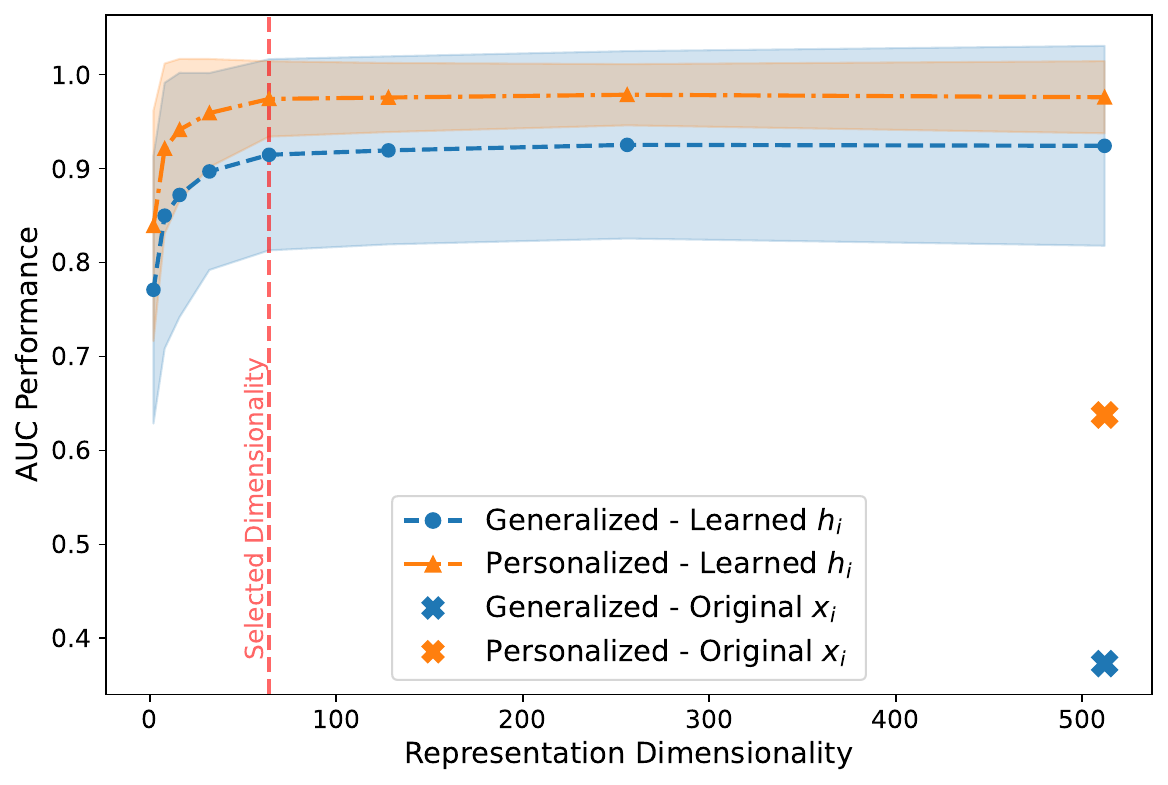} }}%
    \qquad
    \hspace{0.5mm}
    \subfigure[\centering]{{\includegraphics[width=7.3cm, height=5.3cm]{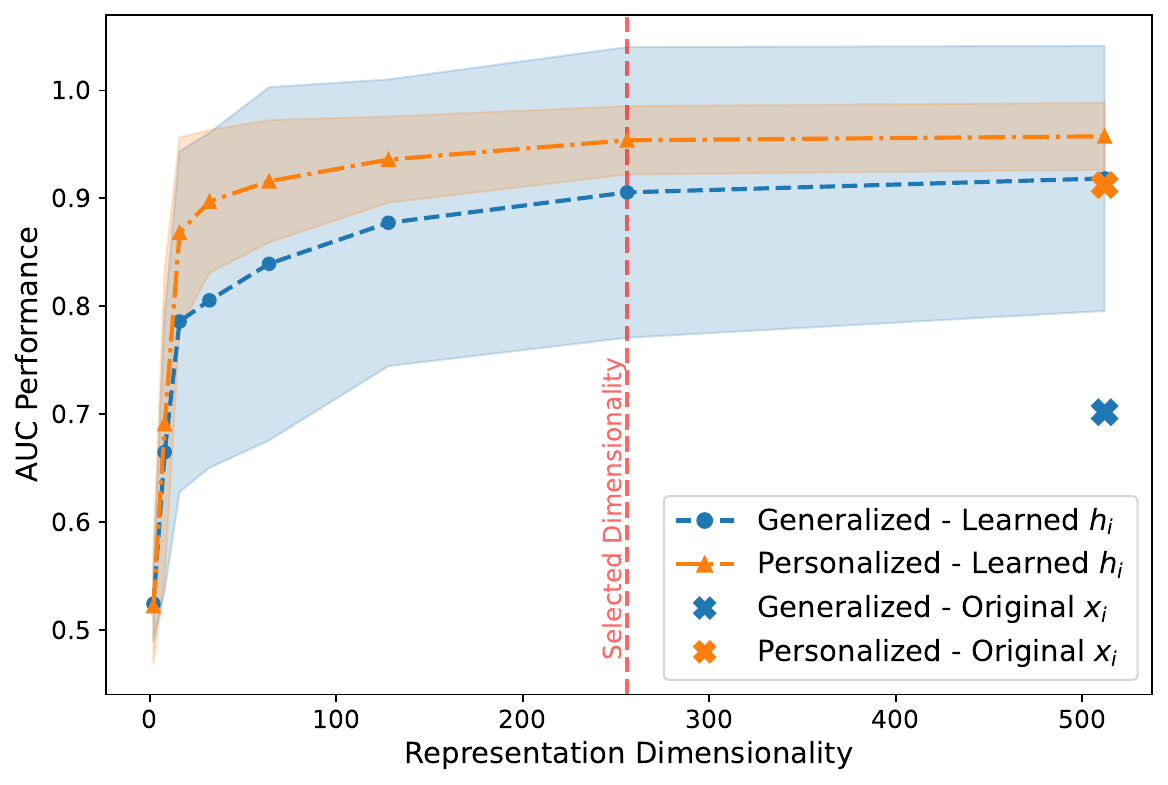} }}%

    \caption{Overview of Movement detection performance in the generalized and personalized scenarios for varying dimensions of the learned representation for a) the PTT-PPG Dataset and b) the Dalia Dataset. The crosses indicate the performance obtained with the original representation.
    }%
    \label{fig5_RepAnalysis}
\end{figure}

%% file: Tables/TableA.tex
\begin{table}[ht]
\renewcommand{\thetable}{A.1} 
\centering
\caption{The detailed architecture of implemented Encoder-Classifier Deep Learning model }
\label{tab:tableA}
\begin{tabular}{p{10cm}|P{2.5cm}|P{1.5cm}}
\toprule
\textbf{Layer (Type)}                               & \textbf{Output Shape} & \textbf{Param \#} \\ \midrule
\multicolumn{3}{l}{\textbf{Encoder}} \\ \midrule
Input Layer                                           & (None, 512, 1)                              & 0                                      \\ 
Conv1D Layer (kernel-size = 64)                        & (None, 449, 32)                             & 2080                                   \\
Conv1D Layer (kernel-size = 64, padding =   “same”)    & (None, 449, 32)                             & 65568                                      \\
Activation Layer                                       & (None, 449, 32)                             & 0                                    \\
Conv1D Layer (kernel-size = 64, padding =   “same”)    & (None, 449, 32)                             & 65568                                      \\
Activation Layer                                       & (None, 449, 32)                            & 0                                 \\
Max Pooling1D Layer                                    & (None, 224, 32)                            & 0                                      \\

Conv1D Layer (kernel-size = 64, padding =   “same”)    & (None, 224, 32)                             & 65568                                      \\
Activation Layer                                       & (None, 224, 32)                             & 0                                    \\
Conv1D Layer (kernel-size = 64, padding =   “same”)    & (None, 224, 32)                             & 65568                                      \\
Activation Layer                                       & (None, 224, 32)                            & 0                                 \\
Max Pooling1D Layer                                    & (None, 112, 32)                            & 0  
\\

Conv1D Layer (kernel-size = 64, padding =   “same”)    & (None, 112, 32)                             & 65568                                      \\
Activation Layer                                       & (None, 112, 32)                             & 0                                    \\
Conv1D Layer (kernel-size = 64, padding =   “same”)    & (None, 112, 32)                             & 65568                                      \\
Activation Layer                                       & (None, 112, 32)                            & 0                                 \\
Max Pooling1D Layer                                    & (None, 56, 32)                            & 0  
\\

Conv1D Layer (kernel-size = 64, padding =   “same”)    & (None, 56, 32)                             & 65568                                      \\
Activation Layer                                       & (None, 56, 32)                             & 0                                    \\
Conv1D Layer (kernel-size = 64, padding =   “same”)    & (None, 56, 32)                             & 65568                                      \\
Activation Layer                                       & (None, 56, 32)                            & 0                                 \\
Max Pooling1D Layer                                    & (None, 28, 32)                            & 0  
\\

Conv1D Layer (kernel-size = 64, padding =   “same”)    & (None, 28, 32)                             & 65568                                      \\
Activation Layer                                       & (None, 28, 32)                             & 0                                    \\
Conv1D Layer (kernel-size = 64, padding =   “same”)    & (None, 28, 32)                             & 65568                                      \\
Activation Layer                                       & (None, 28, 32)                            & 0                                 \\
Max Pooling1D Layer                                    & (None, 14, 32)                            & 0  
\\

Flatten                            & (None, 448)                            & 0                                    \\
Fully Connected Layer                                   & (None, 64)                            & 28736                                      \\
\midrule
\multicolumn{3}{l}{\textbf{Classifier}} \\ \midrule

Fully Connected Output Layer - (Activation = “softmax”)                   & (None, 4)                              & 260                                      \\ \midrule
\multicolumn{3}{l}{\begin{tabular}[c]{@{}l@{}}Total Params: 686,756\\ Trainable Params: 686,756 \& Non-Trainable Params: 0\end{tabular}} \\ \bottomrule
\end{tabular}
\vspace{-0mm}
\end{table}

%% file: Main_paper.bbl
\begin{thebibliography}{28}
\providecommand{\natexlab}[1]{#1}
\providecommand{\url}[1]{\texttt{#1}}
\expandafter\ifx\csname urlstyle\endcsname\relax
  \providecommand{\doi}[1]{doi: #1}\else
  \providecommand{\doi}{doi: \begingroup \urlstyle{rm}\Url}\fi

\bibitem[Allen(2007)]{r3}
John Allen.
\newblock Photoplethysmography and its application in clinical physiological
  measurement.
\newblock \emph{Physiological measurement}, 28\penalty0 (3):\penalty0 R1, 2007.

\bibitem[Ghorbani et~al.(2023)Ghorbani, Reinders, and Tax]{r4}
Ramin Ghorbani, Marcel~JT Reinders, and David~MJ Tax.
\newblock Self-supervised ppg representation learning shows high inter-subject
  variability.
\newblock In \emph{Proceedings of the 2023 8th International Conference on
  Machine Learning Technologies}, pages 127--132, 2023.

\bibitem[Fora et~al.(2019)Fora, Al-Hammouri, and Al-Zaben]{r5}
Malak Fora, Sajidah Al-Hammouri, and Awad Al-Zaben.
\newblock Ecg abnormality detection from ppg signal.
\newblock In \emph{2019 IEEE Jordan International Joint Conference on
  Electrical Engineering and Information Technology (JEEIT)}, pages 103--106.
  IEEE, 2019.

\bibitem[Yousefi et~al.(2018)Yousefi, Parak, Tarniceriu, Harju, Yli-Hankala,
  Korhonen, and Vehkaoja]{r6}
Zeinab~Rezaei Yousefi, Jakub Parak, Adrian Tarniceriu, Jarkko Harju, Arvi
  Yli-Hankala, Ilkka Korhonen, and Antti Vehkaoja.
\newblock Atrial fibrillation detection from wrist photoplethysmography data
  using artificial neural networks.
\newblock In \emph{World congress on medical physics and biomedical
  engineering}, volume 2019, pages 399--404, 2018.

\bibitem[Poh et~al.(2018)Poh, Poh, Chan, Wong, Pun, Leung, Wong, Wong, Chu, and
  Siu]{r7}
Ming-Zher Poh, Yukkee~Cheung Poh, Pak-Hei Chan, Chun-Ka Wong, Louise Pun,
  Wangie Wan-Chiu Leung, Yu-Fai Wong, Michelle Man-Ying Wong, Daniel Wai-Sing
  Chu, and Chung-Wah Siu.
\newblock Diagnostic assessment of a deep learning system for detecting atrial
  fibrillation in pulse waveforms.
\newblock \emph{Heart}, 104\penalty0 (23):\penalty0 1921--1928, 2018.

\bibitem[Boukhechba et~al.(2019)Boukhechba, Cai, Wu, and Barnes]{r8}
Mehdi Boukhechba, Lihua Cai, Congyu Wu, and Laura~E Barnes.
\newblock Actippg: using deep neural networks for activity recognition from
  wrist-worn photoplethysmography (ppg) sensors.
\newblock \emph{Smart Health}, 14:\penalty0 100082, 2019.

\bibitem[Kwon et~al.(2019)Kwon, Hong, Choi, Lee, Hostallero, Kang, Lee, Jeong,
  Koo, Oh, et~al.]{r9}
Soonil Kwon, Joonki Hong, Eue-Keun Choi, Euijae Lee, David~Earl Hostallero,
  Wan~Ju Kang, Byunghwan Lee, Eui-Rim Jeong, Bon-Kwon Koo, Seil Oh, et~al.
\newblock Deep learning approaches to detect atrial fibrillation using
  photoplethysmographic signals: algorithms development study.
\newblock \emph{JMIR mHealth and uHealth}, 7\penalty0 (6):\penalty0 e12770,
  2019.

\bibitem[Chalapathy and Chawla(2019)]{r9_added_1}
Raghavendra Chalapathy and Sanjay Chawla.
\newblock Deep learning for anomaly detection: A survey.
\newblock \emph{arXiv preprint arXiv:1901.03407}, 2019.

\bibitem[Chandola et~al.(2009)Chandola, Banerjee, and Kumar]{r9_added_2}
Varun Chandola, Arindam Banerjee, and Vipin Kumar.
\newblock Anomaly detection: A survey.
\newblock \emph{ACM Comput. Surv.}, 41\penalty0 (3), jul 2009.
\newblock ISSN 0360-0300.
\newblock \doi{10.1145/1541880.1541882}.
\newblock URL \url{https://doi.org/10.1145/1541880.1541882}.

\bibitem[Elgendi et~al.(2019)Elgendi, Fletcher, Liang, Howard, Lovell, Abbott,
  Lim, and Ward]{r4_added_1}
Mohamed Elgendi, Richard Fletcher, Yongbo Liang, Newton Howard, Nigel~H Lovell,
  Derek Abbott, Kenneth Lim, and Rabab Ward.
\newblock The use of photoplethysmography for assessing hypertension.
\newblock \emph{NPJ digital medicine}, 2\penalty0 (1):\penalty0 60, 2019.

\bibitem[Kotorov et~al.(2020)Kotorov, Chi, Shen, et~al.]{r4_added_2}
Rado Kotorov, Lianhua Chi, Min Shen, et~al.
\newblock Personalized monitoring model for electrocardiogram signals:
  diagnostic accuracy study.
\newblock \emph{JMIR Biomedical Engineering}, 5\penalty0 (1):\penalty0 e24388,
  2020.

\bibitem[Bengio et~al.(2013)Bengio, Courville, and Vincent]{r9_added_3}
Yoshua Bengio, Aaron Courville, and Pascal Vincent.
\newblock Representation learning: A review and new perspectives.
\newblock \emph{IEEE transactions on pattern analysis and machine
  intelligence}, 35\penalty0 (8):\penalty0 1798--1828, 2013.

\bibitem[Kramer(1991)]{r10}
Mark~A Kramer.
\newblock Nonlinear principal component analysis using autoassociative neural
  networks.
\newblock \emph{AIChE journal}, 37\penalty0 (2):\penalty0 233--243, 1991.

\bibitem[Chen et~al.(2020)Chen, Kornblith, Norouzi, and Hinton]{r11}
Ting Chen, Simon Kornblith, Mohammad Norouzi, and Geoffrey Hinton.
\newblock A simple framework for contrastive learning of visual
  representations.
\newblock In \emph{International conference on machine learning}, pages
  1597--1607. PMLR, 2020.

\bibitem[Zhang et~al.(2023)Zhang, Geng, and Hong]{r12}
Wenrui Zhang, Shijia Geng, and Shenda Hong.
\newblock A simple self-supervised ecg representation learning method via
  manipulated temporal--spatial reverse detection.
\newblock \emph{Biomedical Signal Processing and Control}, 79:\penalty0 104194,
  2023.

\bibitem[Bozorgtabar et~al.(2020)Bozorgtabar, Mahapatra, Vray, and Thiran]{r13}
Behzad Bozorgtabar, Dwarikanath Mahapatra, Guillaume Vray, and Jean-Philippe
  Thiran.
\newblock Salad: Self-supervised aggregation learning for anomaly detection on
  x-rays.
\newblock In \emph{International Conference on Medical Image Computing and
  Computer-Assisted Intervention}, pages 468--478. Springer, 2020.

\bibitem[Venkatakrishnan et~al.(2020)Venkatakrishnan, Kim, Eisawy, Pfister, and
  Navab]{r14}
Abinav~Ravi Venkatakrishnan, Seong~Tae Kim, Rami Eisawy, Franz Pfister, and
  Nassir Navab.
\newblock Self-supervised out-of-distribution detection in brain ct scans.
\newblock \emph{arXiv preprint arXiv:2011.05428}, 2020.

\bibitem[Zhao et~al.(2021)Zhao, Li, He, Ma, Fang, Li, and Zheng]{r15}
He~Zhao, Yuexiang Li, Nanjun He, Kai Ma, Leyuan Fang, Huiqi Li, and Yefeng
  Zheng.
\newblock Anomaly detection for medical images using self-supervised and
  translation-consistent features.
\newblock \emph{IEEE Transactions on Medical Imaging}, 40\penalty0
  (12):\penalty0 3641--3651, 2021.

\bibitem[Li et~al.(2021)Li, Sohn, Yoon, and Pfister]{r16}
Chun-Liang Li, Kihyuk Sohn, Jinsung Yoon, and Tomas Pfister.
\newblock Cutpaste: Self-supervised learning for anomaly detection and
  localization.
\newblock In \emph{Proceedings of the IEEE/CVF Conference on Computer Vision
  and Pattern Recognition}, pages 9664--9674, 2021.

\bibitem[Ristea et~al.(2022)Ristea, Madan, Ionescu, Nasrollahi, Khan, Moeslund,
  and Shah]{r17}
Nicolae-C{\u{a}}t{\u{a}}lin Ristea, Neelu Madan, Radu~Tudor Ionescu, Kamal
  Nasrollahi, Fahad~Shahbaz Khan, Thomas~B Moeslund, and Mubarak Shah.
\newblock Self-supervised predictive convolutional attentive block for anomaly
  detection.
\newblock In \emph{Proceedings of the IEEE/CVF Conference on Computer Vision
  and Pattern Recognition}, pages 13576--13586, 2022.

\bibitem[Xu et~al.(2020)Xu, Zheng, Mao, Wang, and Zheng]{r18}
Junjie Xu, Yaojia Zheng, Yifan Mao, Ruixuan Wang, and Wei-Shi Zheng.
\newblock Anomaly detection on electroencephalography with self-supervised
  learning.
\newblock In \emph{2020 IEEE International Conference on Bioinformatics and
  Biomedicine (BIBM)}, pages 363--368. IEEE, 2020.

\bibitem[Zhang et~al.(2022)Zhang, Wang, Chen, Yu, and Qin]{r19}
Yuxin Zhang, Jindong Wang, Yiqiang Chen, Han Yu, and Tao Qin.
\newblock Adaptive memory networks with self-supervised learning for
  unsupervised anomaly detection.
\newblock \emph{IEEE Transactions on Knowledge and Data Engineering}, 2022.

\bibitem[Tax(2002)]{r19_added_1}
David Martinus~Johannes Tax.
\newblock One-class classification: Concept learning in the absence of
  counter-examples.
\newblock 2002.

\bibitem[Liu et~al.(2008)Liu, Ting, and Zhou]{r19_added_2}
Fei~Tony Liu, Kai~Ming Ting, and Zhi-Hua Zhou.
\newblock Isolation forest.
\newblock In \emph{2008 eighth ieee international conference on data mining},
  pages 413--422. IEEE, 2008.

\bibitem[Mehrgardt et~al.(2022)Mehrgardt, Khushi, Poon, and Withana]{r20}
Philip Mehrgardt, Matloob Khushi, Simon Poon, and Anusha Withana.
\newblock Pulse transit time {PPG} dataset, 2022.

\bibitem[Reiss et~al.(2019)Reiss, Indlekofer, Schmidt, and Van~Laerhoven]{r21}
Attila Reiss, Ina Indlekofer, Philip Schmidt, and Kristof Van~Laerhoven.
\newblock Deep ppg: large-scale heart rate estimation with convolutional neural
  networks.
\newblock \emph{Sensors}, 19\penalty0 (14):\penalty0 3079, 2019.

\bibitem[Zhang(2015)]{r22}
Zhilin Zhang.
\newblock Photoplethysmography-based heart rate monitoring in physical
  activities via joint sparse spectrum reconstruction.
\newblock \emph{IEEE transactions on biomedical engineering}, 62\penalty0
  (8):\penalty0 1902--1910, 2015.

\bibitem[Salehizadeh et~al.(2015)Salehizadeh, Dao, Bolkhovsky, Cho, Mendelson,
  and Chon]{r23}
Seyed~MA Salehizadeh, Duy Dao, Jeffrey Bolkhovsky, Chae Cho, Yitzhak Mendelson,
  and Ki~H Chon.
\newblock A novel time-varying spectral filtering algorithm for reconstruction
  of motion artifact corrupted heart rate signals during intense physical
  activities using a wearable photoplethysmogram sensor.
\newblock \emph{Sensors}, 16\penalty0 (1):\penalty0 10, 2015.

\end{thebibliography}
